\RequirePackage{fix-cm}
\documentclass[twocolumn]{svjour3}          
\smartqed  
\usepackage{graphicx}
\usepackage{amsmath} 
\usepackage{amssymb} 
\usepackage{sidecap} 
\sidecaptionvpos{figure}{t}
\usepackage{caption}
\usepackage{lipsum}

\newcommand{\etal}{\textit{et al}. }

\journalname{}
\begin{document}\sloppy

\title{AnimalWeb: A Large-Scale Hierarchical Dataset of Annotated Animal Faces
}


\author{Muhammad Haris Khan         \and
        John McDonagh          \and
        Salman Khan   \and
        Muhammad Shahabuddin \and
        Aditya Arora \and
        Fahad Shahbaz Khan \and
        Ling Shao \and
        Georgios Tzimiropoulos
}


\institute{M. Haris Khan \at Inception Institute of Artificial Intelligence, Abu-Dhabi, UAE \\
             \email{muhammad.haris@inceptioniai.org}           
           \and
           John McDonagh \at University of Nottingham, UK \\
           \email{john.mcdonagh@nottingham.ac.uk}
           \and
           Salman Khan \at Inception Institute of Artificial Intelligence, Abu-Dhabi, UAE \\
           \email{salman.khan@inceptioniai.org}
           \and
           M. Shahabuddin \at Comsats University Islamabad, Pakistan \\
           \email{shahab.pk05@gmail.com}
           \and
           Aditya Arora \at Inception Institute of Artificial Intelligence, Abu-Dhabi, UAE \\
           \email{aditya.arora@inceptioniai.org}
           \and
           Fahad S. Khan \at Inception Institute of Artificial Intelligence, Abu-Dhabi, UAE \\
           \email{fahad.khan@inceptioniai.org}
           \and
           Ling Shao \at Inception Institute of Artificial Intelligence, Abu-Dhabi, UAE \\
           \email{ling.shao@inceptioniai.org}
           \and
           Georgios Tzimiropoulos \at University of Nottingham, UK \\
           \email{Yorgos.Tzimiropoulos@nottingham.ac.uk}
}

\date{Received: date / Accepted: date}

\maketitle

\begin{abstract}
Our existence on this planet is heavily reliant on animals. It is our ethical obligation to improve their well-being by understanding their needs. Several studies show that animal needs are often expressed through their faces and mammalian brains are capable enough to decode social signals from fellow animal faces. Though remarkable progress has been made towards the automatic understanding of human faces, this has regrettably not been the case with animal faces. There exists significant room and appropriate need to develop automatic systems capable of interpreting animal faces. Among many transformative impacts, such a technology will foster better and cheaper animal healthcare, and further advance animal psychology understanding.

We believe the underlying research progress is mainly obstructed by the lack of an adequately annotated dataset of animal faces, covering a wide spectrum of animal species. To this end, we introduce a large-scale, hierarchical annotated dataset of animal faces, featuring 21.9K faces captured `in-the-wild' conditions. These faces belong to 334 diverse species, while covering 21 different animal orders across biological taxonomy. Each face is consistently annotated with 9 landmarks on key facial features. It is structured and scalable by design; its development underwent four systematic stages involving rigorous, manual annotation effort of over 6K man-hours. We benchmark the proposed dataset for face alignment using the existing art under two new problem settings. Results showcase its challenging nature, unique attributes and present definite prospects for novel, adaptive, and generalized face-oriented CV algorithms. We further benchmark the dataset across related tasks, namely face detection and fine-grained recognition, to demonstrate multi-task applications and opportunities for improvement. Experimental evaluation indicates that this dataset will push the algorithmic advancements across many related CV tasks and encourage the development of novel systems for animal facial behaviour monitoring. We will make the dataset publicly available.



\end{abstract}

\section{Introduction}
\label{intro}
Animals are a fundamental part of our world. It is our moral duty to improve the condition and well-being of animals in labs, farms and homes by understanding their needs and requirements often expressed through their faces. Behavioural and neurophysiological literature have shown that mammalian brains can interpret social signals on fellow animal’s faces and have developed specialized skills to process facial features. Therefore, the study of animal faces is of prime importance. 
\begin{figure}[t]
\begin{center}
    \includegraphics[width=\linewidth]{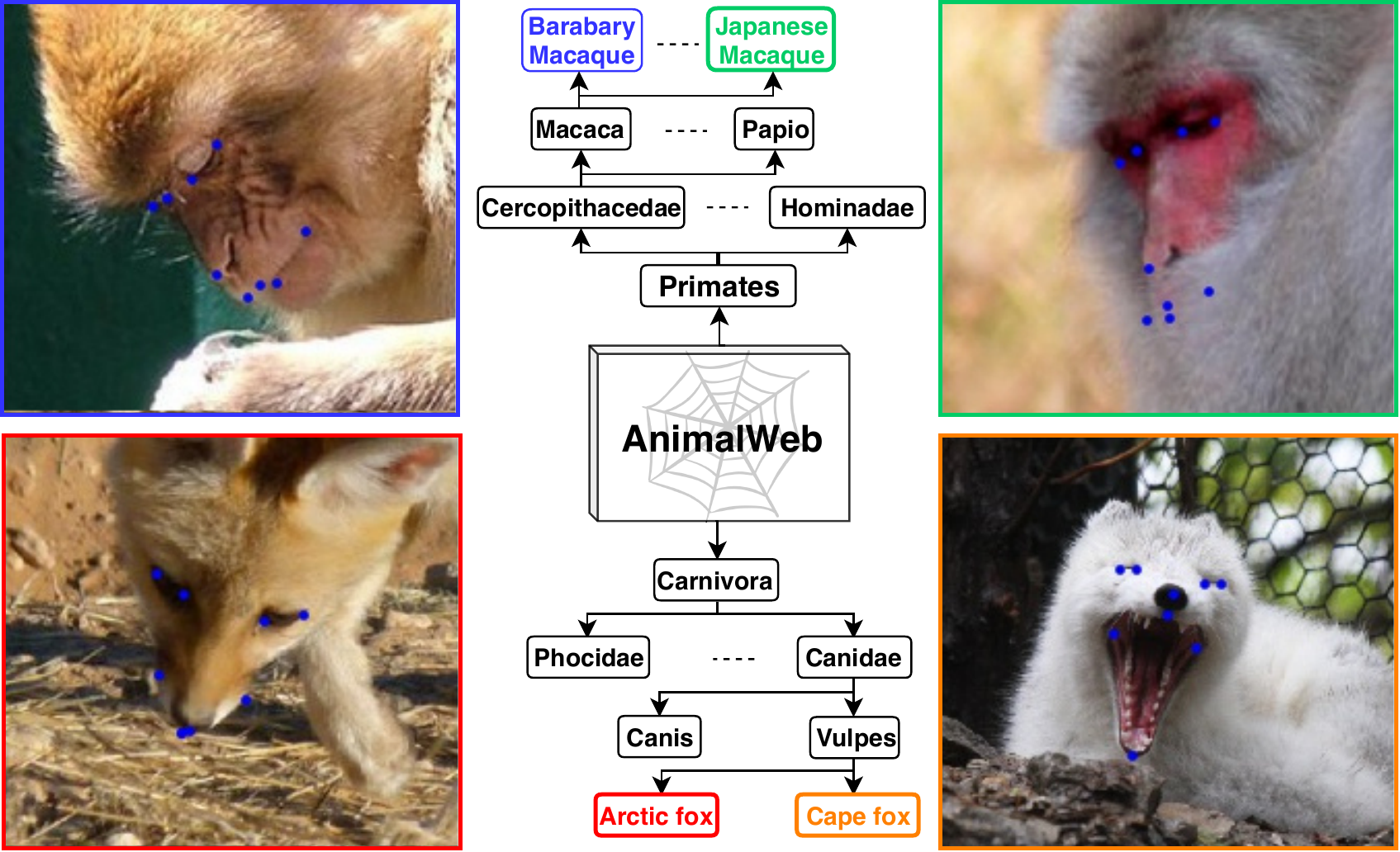}
\end{center} 
   \caption{\small \emph{AnimalWeb:} We introduce a large-scale, hierarchical dataset of annotated animal faces featuring diverse species while covering a broader spectrum of animal biological taxonomy. The dataset exhibits unique challenges e.g., large biodiversity in species, high variations in pose, scale, appearance, deformations and backgrounds. Further, it offers unique attributes like class imbalance (CI), multi-task applications (MTA), and zero-shot face alignment (ZFA). Facial landmarks shown in blue and the images belong to classes with identical color in the hierarchy. 
   }
   \label{fig:teaser_fig}
\end{figure}


Facial landmarks can help us better understand animals and foster their well-being via deciphering their facial expressions. Facial expressions reflect the internal emotions and psychological state of an animal being. As an example, animals with different anatomical structure (such as mice, horses, rabbits and sheep), show a similar grimace expression when in pain i.e., tighten eyes and mouth, flatten cheeks and unusual ear postures. Understanding abnormal animal expressions and behaviours with visual imagery is a much cheaper and quicker alternative to clinical examinations and vital signs monitoring.  
Encouraging indicators show such powerful technologies could indeed be possible, e.g., fearful cows widen their eyes and flatten their ears \cite{kutzer2015habituation}, horses close eyes in depression \cite{fureix2012towards}, sheep positions its ears backward when facing unpleasant situations \cite{boissy2011cognitive}, and rats ear change colors and shape when in joy \cite{finlayson2016facial}. Furthermore, large-scale annotated datasets of animal faces can help advance the animal psychology understanding to a new level. For example, for non-primate animals, the scientific understanding of animal expressions is generally limited to the development of only pain coding systems. However, other expressions could be equally important to understand e.g., sadness, boredom, hunger, anger and fear.

We believe the research progress towards automatic understanding of animal facial behaviour is largely hindered by the lack of sufficiently annotated animal faces, covering a wide spectrum of animal species. In comparison, significant progress has been made towards automatic understanding and interpretation of human faces \cite{xiong2013supervised,cao2014face,tzimiropoulos2015project,trigeorgis2016mnemonic,bulat2017far,masi2016we,wang2017face}, while animal face analysis is largely unexplored in vision community \cite{yang2016human,rashid2017interspecies}. There is a plenty of room for new algorithms and a pressing need to develop computational tools capable of understanding animal facial behavior. 
 To this end, we introduce a large-scale, hierarchical dataset of annotated animal faces, termed AnimalWeb, featuring 
 diverse species while covering a broader spectrum of animal biological taxonomy. Fig.~\ref{fig:teaser_fig} provides a holistic overview of the dataset key features. 
 

AnimalWeb construction follows the well established hierarchy of animals biological classification. In animal kingdom, the tree begins from Phylum and boils down to Class, Order, Family, Genus, and Species. Every image in the dataset has been labelled with the genus-species i.e. the leaf of this classification tree. Image collection is driven by the motivation to offer complete in-the-wild conditions (such as pose, expression, illumination, and occlusions) and diverse coverage of orders in the animal kingdom. 


\textbf{Contributions:} To our knowledge, we build and annotate the largest dataset of animal faces captured under altogether in-the-wild conditions. It encompasses 21 different orders across animal biological taxonomy. Each order probes various families (ranging from 1 to 12), and each family further explores an average of 8 genuses. This diverse coverage makes up a total of 334 different animal species resulting in a count of 21.9K animal faces. Each face is consistently annotated with 9 fiducial landmarks centered around key facial components such as eyes and mouth. Finally, the dataset design and development followed four systematic stages involving an overall, rigorous, manual labelling effort of 6,833 man-hours by experts and trained volunteers. 

We benchmark AnimalWeb for face alignment with the state-of-the-art human face alignment algorithms \cite{bulat2017far,xiong2017combining}. Results indicate that the dataset is challenging for current best methods developed for human face alignment particularly due to biodiversity, specie imbalance, and adverse in-the-wild conditions (e.g., extreme poses). We show results under two different settings, namely known species evaluation and unknown species evaluation. These settings reveal the capability of the proposed dataset for testing under two novel problem settings: few-shot face alignment and zero-shot face alignment. Further, we demonstrate related applications possible with this dataset, in particular, animal face detection and fine-grained specie recognition. Experimental results signal that the dataset is a strong experimental base for algorithmic advances in computer vision. For instance, the development of novel, adaptive, and generalized facial alignment algorithms towards the betterment of society and economy.

\begin{figure*}[t]
\begin{center}
    \includegraphics[width=\linewidth]{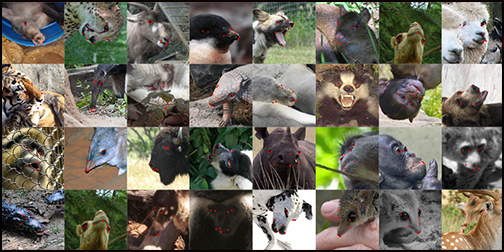}
\end{center} 
   \caption{\small Some representative examples from randomly chosen species in AnimalWeb. Animal faces tend to exhibit large variations in pose, scale, appearance and expressions.} 
   \label{fig:main_fig}
\end{figure*}

\section{Related Datasets}
Owing to ever-growing interest in automatic face analysis, several face alignment datasets mainly targeting human faces have been published \cite{gross2010multi,sagonas2013300,shen2015first,deng2018menpo}. However, there has been little to no progress towards creating datasets for animal faces at a comparable scale \cite{yang2016human,rashid2017interspecies}. In this section, we categorize existing human and animal face alignment benchmarks according to their level of difficulty and briefly overview each category.

\subsection{Human Face Alignment}
\noindent \textbf{Low Difficulty Datasets:} Since the seminal work of Active Appearance Models (AAMs) \cite{cootes1998active}, various 2D datasets featuring human face landmark annotations have been proposed. Among these, the prominent ones are XM2VTS \cite{messer1999xm2vtsdb}, BioID \cite{jesorsky2001robust}, FRGC \cite{phillips2005overview}, and Multi-PIE \cite{gross2010multi}. These datasets were collected under constrained environments with limited expression, frontal pose, and normal lighting variations. Following them, few datasets were proposed with faces showing occlusions and other variations such as COFW \cite{burgos2013robust,ghiasi2015occlusion} and AFW \cite{zhu2016face}.


\noindent \textbf{Moderate Difficulty Datasets:} 300W \cite{sagonas2013300} is considered a popular dataset amongst several others in human face alignment. It has been widely adopted both by scientific community as well as industry \cite{trigeorgis2016mnemonic,xiong2013supervised,ren2014face,zhu2016unconstrained}. This benchmark was developed for the 300W competition held in conjunction with ICCV 2013. 300W benchmark originated from LFPW \cite{belhumeur2013localizing}, AFW \cite{zhu2016face}, IBUG \cite{sagonas2013300}, and 300W private \cite{sagonas2016300} datasets. In total, it provides 4,350 images with faces annotated using the 68 landmark frontal face markup scheme. In pursuit of promoting face tracking research, 300VW \cite{shen2015first} is introduced featuring 114 videos. Such datasets paced research progress towards human face alignment in challenging conditions. 



\noindent \textbf{High Difficulty Datasets:} More recently, efforts are directed to manifest greater range of variations. For instance, Annotated Facial Landmarks in the wild (AFLW) \cite{koestinger2011annotated} proposed a collection of 25K annotated human faces with up to 21 landmarks. It, however, excluded locations of invisible landmarks. Zhu \etal \cite{zhu2016unconstrained} provided manual annotations for invisible landmarks, but there are no landmark annotations along the face contour. Along similar lines, Zhu \etal \cite{zhu2016face} developed a large scale training dataset by synthesizing profile views from 300W dataset using a 3D Morphable Model (3DMM). Though it could serve as a large training set, the synthesized profile faces have artifacts that can hurt fitting accuracy. Jeni \etal \cite{jeni2016first} reported a dataset introduced in a competition held along ECCV 2016; it typically consisted of images photographed in controlled conditions or are produced synthetically.

Lately, Menpo benchmark \cite{deng2018menpo} was released as part of competitions held along ICCV 2017. It contains landmarks annotations both from 2D and 3D perspectives and exhibits large variations in pose, expression, illumination and occlusions. Faces are also classified into semi-frontal and profile based on their orientation and annotated accordingly. Menpo 2D benchmark contains 7,576 and 7,281 annotated training and testing images, respectively, taken from AFLW and FDDB. 

\subsection{Animal Face Alignment}
Despite scientific value, pressing need and direct impact on animal health and welfare, only little attention has been paid in developing an annotated dataset of animal faces \cite{yang2016human,rashid2017interspecies}. Although datasets such as ImageNet \cite{deng2018menpo} and iNaturalist \cite{van2018inaturalist} offer reasonable species variety, they are targeted at image-level classification and region-level detection tasks. The two animal face alignment datasets are reported in \cite{yang2016human} and \cite{rashid2017interspecies}. Yang \etal \cite{yang2016human} collected 600 sheep faces and annotated them with 8 fiducial landmarks. Similarly, Rashid \etal \cite{rashid2017interspecies} reported a collection of 3717 horse faces with points marked around 8 facial features. These datasets are severely limited in terms of biodiversity, size, and range of possible real-world conditions. To the best of our knowledge, the proposed dataset is a first large-scale, hierarchical collection of annotated animal faces with 9 landmarks. It possess real-world properties e.g., large variations in pose, scale and appearance as well as unique attributes such as species imbalance, multi-task applications, and zero-shot face alignment. Next, we introduce our proposed dataset.

\section{Dataset Properties}

AnimalWeb has been constructed following the animal biological taxonomy. It populates faces from 334 different species spread over 21 different animal orders. Below, we highlight some of the unique aspects of this newly introduced dataset (Fig.~\ref{fig:main_fig}).  


\begin{table}[htp]
\begin{center}
\resizebox{\linewidth}{!}{
\begin{tabular}{lccc}
\hline
\textbf{Dataset}    & \textbf{Target Face}  & \textbf{Faces} & \textbf{Points}    \\ \hline
Multi-PIE \cite{gross2010multi} (semi-frontal)         &Human &6665  &68   \\
Multi-PIE \cite{gross2010multi} (profile)   &Human     &1400  &39   \\
AFLW \cite{koestinger2011annotated}  &Human &25,993 &21    \\
COFW \cite{burgos2013robust}             &Human     &1007  &29 \\
COFW \cite{ghiasi2015occlusion}          &Human     &507  &68  \\
300 W\cite{sagonas2013300,sagonas2016300}    &Human &3837    & 68  \\
Menpo 2D \cite{deng2018menpo} (semi-frontal)   &Human &10,993    & 68  \\
Menpo 2D \cite{deng2018menpo} (profile)      &Human    &3852   & 39 \\
AFLW2000-3D \cite{zhu2016face}           &Human        &2000   & 68 \\
300W-LP \cite{zhu2016face}(synthetic)    &Human        &61,225   & 68 \\ \hline
Sheep faces \cite{yang2016human}        &Animal       &600   & 8 \\
Horse faces \cite{rashid2017interspecies} &Animal     &3717   & 8 \\  
AnimalWeb (Ours)                        &Animal      &21,921   & 9 \\
\hline
\end{tabular} 
}
\end{center} 
\caption{\small Comparison between AnimalWeb and various popular face alignment datasets. We see that AnimalWeb is bigger (in terms of faces offered) than 80\% of the datasets targeted at human face alignment. Further, the existing efforts on animal face datasets are limited to only single species. This work targets a big gap in this area and builds a large-scale annotated animal faces dataset. It possess real-world properties and exhibits unique attributes like class imbalance (CI), multi-task applications (MTA), and zero-shot face alignment (ZFA) as shown in experiments.}

\label{tab:comp_datasets}
\end{table}

\begin{figure}
\centering
    \includegraphics[width=0.8\linewidth]{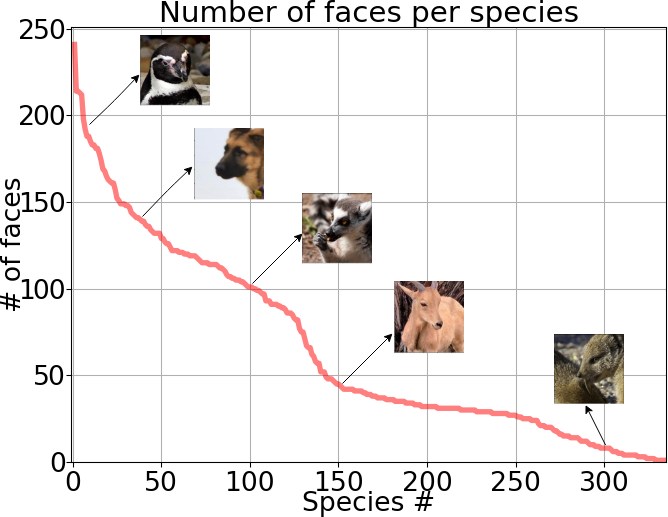}
   \hspace{-1.0em}
   \caption{\small Distribution of faces per specie in AnimalWeb. We see that 29\% of the total species contain 65\% of the total faces. The dataset shows the natural occurrence patterns of different species.}
   \label{fig:species_dist}
\end{figure}

\noindent\textbf{Scale:}
The proposed dataset is aimed at offering a large-scale and diverse coverage of annotated animal faces. It contains 21.9K annotated faces, offering 334 different animal species with variable number of animal faces in each species. Fig.~\ref{fig:species_dist} shows the distribution of faces per specie. We see that 29\% of the total species contain 65\% of the total faces. Also, the maximum and minimum number of faces per specie are 241 and 1, respectively. Both these statistics highlight the large imbalance between species and high variability in the instance count for different species. This marks the conformity with the real-world where different species are observed with varying frequencies.

Offered species in AnimalWeb cover 21 different orders from animal classification tree. An average of 3 families have been covered in each order. Similarly, on average 8 genuses have been explored in each family. To the best of our knowledge, AnimalWeb is the first large-scale dataset of annotated animal faces that is easily scalable to offer greater biodiversity coverage in a principled way. It can be highly impactful, for instance, annotated faces could play a vital role in interpreting greater variety of animal expressions not possible with the current approaches based solely on pain coding systems. Tab.~\ref{tab:comp_datasets} draws a comparison between AnimalWeb and various popular datasets for face alignment. We see that AnimalWeb is bigger (in face count) compared to 80\% of datasets targeted at human face alignment. Importantly, very little or rather no attention is subjected towards constructing annotated animal faces dataset mimicking real-world properties, and the existing ones are limited to only single species.


\noindent \textbf{Diversity:}
Robust computational tools aimed at detecting/tracking animal facial behaviour in open environments are difficult to realize without observations that can exhibit real-world scenarios as much as possible. We therefore aim at ensuring diversity along two important dimensions, \textbf{(1)} imaging variations in scale, pose, expression, and occlusion, \textbf{(2)} species coverage in the animal biological taxonomy. Fig.~\ref{fig:main_fig} shows some example variations captured in the dataset. We observe that animal faces exhibit great pose variations and their faces are captured from very different angles (e.g., top view) that are quite unlikely for human faces. In addition, animal faces can show great range of pose and scale variations. 

\begin{figure}[h]
\begin{center}
    \includegraphics[width=\linewidth]{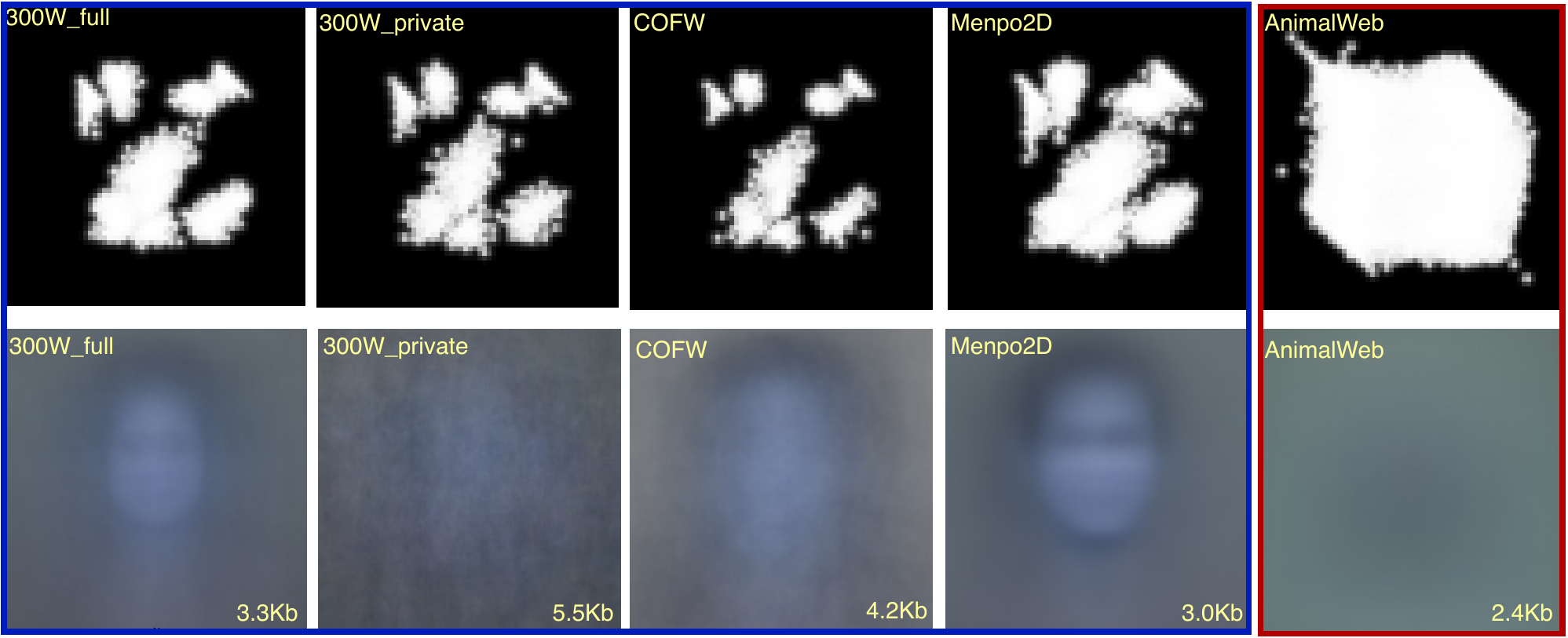}
\end{center} 
   \caption{\small Top: AnimalWeb covers significantly larger space of deformations compared to popular human face alignment datasets.  
   Bottom: It offers more diversity - large variability in appearances, viewpoints, poses, clutter and occlusions resulting in the blurriest mean image with the smallest lossless JPG file size when compared to popular human face alignment datasets.} 
   \label{fig:pose_diversity}
\end{figure}

Fig.~\ref{fig:pose_diversity} (top row) reveals that faces in AnimalWeb exhibits much greater range of shape deformations. Each image is obtained by warping all possible ground truth shapes to a reference shape, thereby removing similarity transformations. The bottom row in Fig.~\ref{fig:pose_diversity} attempts to demonstrate image diversification in AnimalWeb and other datasets. We observe that it comprises more diversified images than other commonly available human face alignment datasets.

\begin{figure}[]
\centering
    \includegraphics[width=0.8\linewidth]{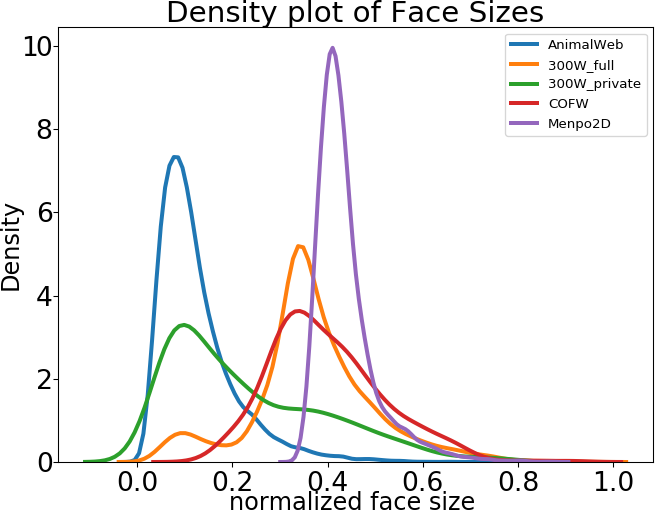}
   \hspace{-1em}
   \caption{\small Face sizes distribution in AnimalWeb and popular human face alignment datasets. AnimalWeb offers 32\% more range of small face sizes ($<0.2$) in comparison to competing datasets. Face sizes along x-axis are normalized by images size.} 
   \label{fig:face_dist}
\end{figure}

To gauge scale diversity, we plot the distribution of normalized face sizes for AnimalWeb in Fig.~\ref{fig:face_dist} and popular human face alignment datasets. AnimalWeb offers 32\% more range of small face sizes ($<0.2$) in comparison to competing datasets for human face alignment.



\begin{figure}[!htp]
\begin{center}
    \includegraphics[width=\linewidth]{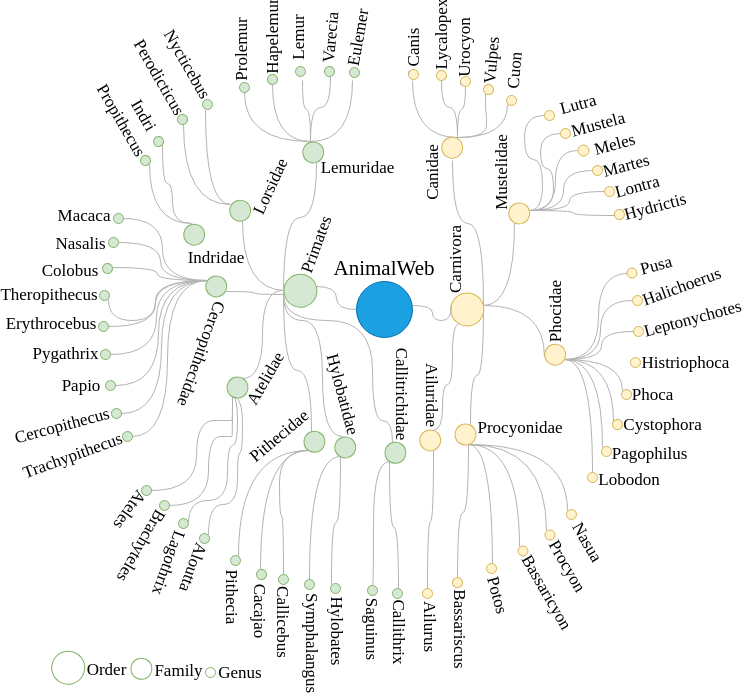}
\end{center} 
   \caption{\small A miniature glimpse of the hierarchical nature of AnimalWeb. Two different orders, Primates and Carnivora, have been shown with 8 and 5 families along with some of their respective genuses.} 
   \label{fig:species_miniature}
\end{figure}

Fig.~\ref{fig:species_miniature} provides a miniature view of the hierarchical nature, illustrating diversity of the dataset. Two different orders, Primates and Carnivora, have been shown with randomly chosen 8 and 5 families along with some of their respective genuses. It can be seen that AnimalWeb exhibits hierarchical structure with variable number of children nodes for each parent node. We refer to Tab.~\ref{tab:dist_families} for the count of families, genuses, species, and finally faces in every order present in the dataset. There exists a total of 21 orders and each order explores on average 3 families, 8 genuses, and 1024 faces. Primates and Carnivora orders populate maximum number of families i.e. 12 among others. We see a similar trend further down the hierarchy. Both aforementioned orders also comprise maximum count of genuses, species, and faces. 

\section{Constructing AnimalWeb}
\begin{figure*}[t]
    \centering
    \includegraphics[width=\linewidth, clip=true, trim=0.8cm 2cm 1.5cm 3cm]{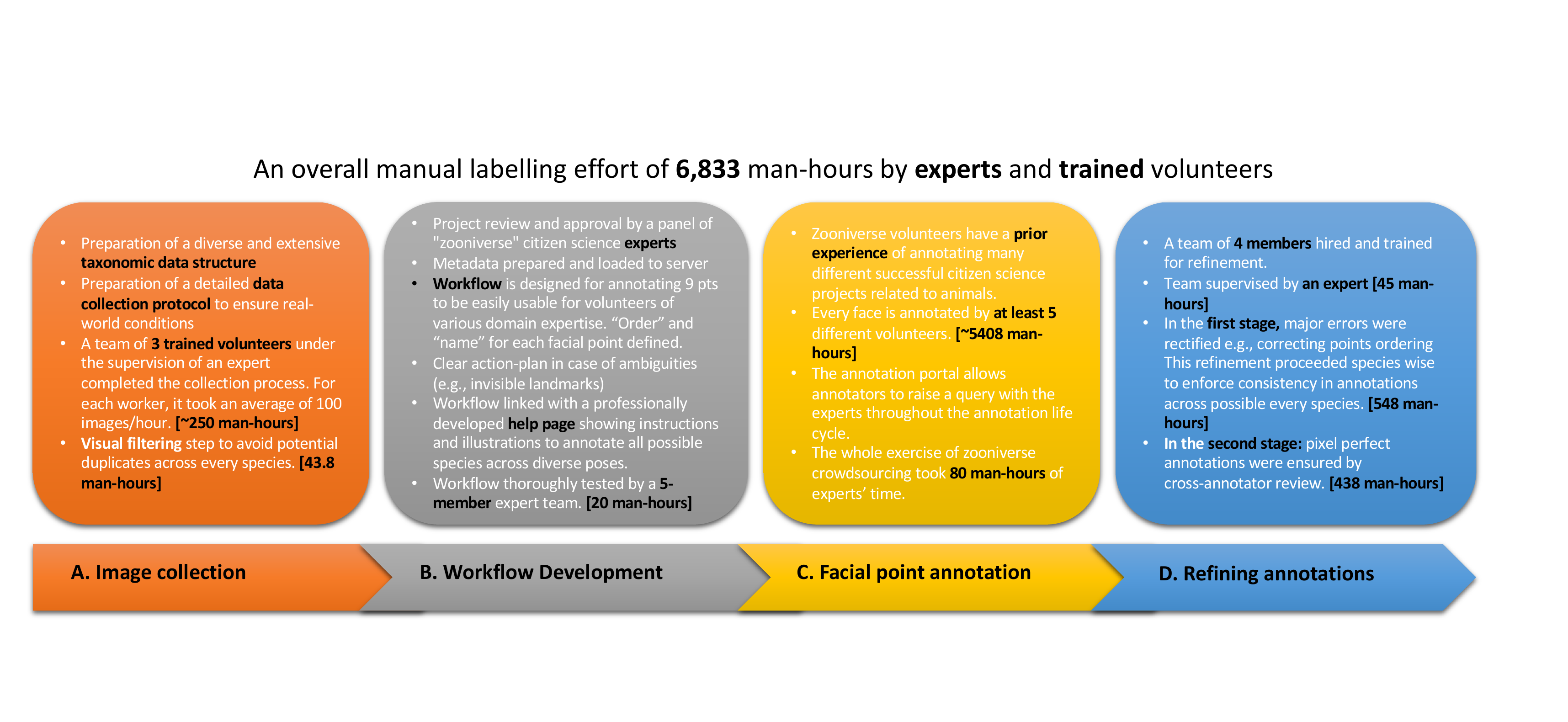}
    \caption{Four systematic stages in AnimalWeb development with associated details and man-hours involved. Zoom-in for details.}
    \label{fig:dataset_lifecycle}
\end{figure*}

In this section, we detail four important steps followed towards the construction of the proposed dataset (see Fig.~\ref{fig:dataset_lifecycle}). These steps include image collection, workflow development, facial point annotation, and annotation refinement. We elaborate these further below. 

\subsection{Image Collection}
\label{sec:img_collection}
To achieve image collection, we first developed a taxonomic framework to realise a structured, scalable dataset design followed by a detailed collection protocol to ensure real-world conditions before starting image collection process.

\noindent \textbf{Taxonomic Framework Development.} We develop a taxonomic framework for the AnimalWeb dataset. A simple, hierarchical tree-like data structure is designed following the well established biological animal classification. The prime motivation for this is to carry out image collection - the next step in dataset construction - in a structured and principled way. The obvious other advantage for this methodology lies in recording the various statistics such as image count at different nodes of the tree.  

\noindent \textbf{Data Collection Protocol.} Starting from animal kingdom we restricted ourselves to vertebrates group (phylum), and further within vertebrates to Mammalia class. We wanted those animals whose faces exhibit roughly regular and identifiable face structure. Some excluded animal examples are insects and worms that possibly violate this condition. Given these restrictions, 21 orders were shortlisted for collection task, whom scientific names are depicted in Tab.~\ref{tab:dist_families}.


\begin{table}[!htp]
\centering
\resizebox{1\columnwidth}{!}{
\begin{tabular}{lcccc}
\hline
\textbf{Order} & \textbf{Families} & \textbf{Genuses} & \textbf{Species} & \textbf{Faces} \\ \hline
Tubulidentata &1 &1 &1 &34 \\
Carnivora &11 &57 &144 &8281 \\
Artiodactyla &7 &42 &55 &4546 \\
Sphenisciformes &1 &5 &10 &1516 \\
Diprotodontia &3 &7 &14 &775 \\
Rodentia &11 &19 &19 &1521 \\
Lagomorpha &1 &2 &4 &86 \\
Pilosa &1 &1 &1 &48 \\
Cingulata &1 &1 &1 &58 \\
Peramelemorphia &1 &1 &1 &61 \\
Primates &12 &30 &59 &3468 \\
Perissodactyla &2 &3 &10 &930 \\
Crocodilia &2 &2 &2 &168 \\
Sirenia &1 &1 &1 &25 \\
Dasyuromorphia &1 &3 &3 &54 \\
Monotremata &2 &2 &2 &113 \\
Eulipotyphla &1 &1 &1 &32 \\
Hyracoidea &1 &1 &1 &82 \\
Microbiotheria &1 &1 &1 &4 \\
Didelphimorphia &1 &1 &1 &67 \\
Marsupialia &1 &1 &1 &31 \\
\hline
\end{tabular} }
\caption{\small List of orders covered in AnimalWeb and for each order we show the number of families, genuses, species, and faces. There are a total of 21 orders and each order explores on average 3 families, 8 genuses, and 1024 faces.}
\label{tab:dist_families}
\end{table}


Finally, we set the bound for number of images to be collected per genus-species between 200-250. This would increase the chances of valuable collection effort to be spent in exploring the different possible species - improving biodiversity - rather than heavily populating a few (commonly seen). With this constraint, we ended up with an average of 65 animal faces per specie.

\noindent \textbf{Image Source.} The Internet is the only source used for collecting images for this dataset. Other large-scale computer vision datasets such as ImageNet \cite{deng2009imagenet} and MS COCO \cite{lin2014microsoft} have also relied on this source to achieve the same. Specifically, we choose Flickr\textit{\footnote{https://www.flickr.com/}}, which is a large image hosting website, to search first, then select, and finally download relevant animal faces.

\noindent \textbf{Collection.} We use both common and scientific names of animal species from the taxonomic framework (described earlier) to query images. Selection is primarily based on capturing various in-the-wild conditions e.g. various face poses. A team of 3 trained volunteers completed the image collection process under the supervision of an expert. For each worker, it took an average of 100 images per hour amounting to a total of $\sim$250 man-hours. After download, we collected around 25K candidate images. Finally, a visual filtering step helped removing potential duplicates across species in 43.8 man-hours.


\subsection{Workflow Development}
\label{subsection:workflow_dev}
Annotating faces can be regarded as the most important, labour-intensive and thus a difficult step towards this dataset construction. To actualize this, we leveraged the great volunteers resource from a large citizen science web portal, called Zooniverse \footnote{https://www.zooniverse.org/}. It is home to many successful citizen science projects. We underwent the following stages to accomplish successful project launch through this portal. 

\noindent \textbf{Project Review.} This is the \emph{first} stage and it involves project design and review. The project is only launched once it gets reviewed by Zooniverse experts panel whom main selection criterion revolves around gauging the impact of a research project. 

\noindent \textbf{Workflow design and development.} Upon clearing review process, in the \emph{second} phase, the relevant image metadata is uploaded to the server and an annotator interface (a.k.a workflow) is developed. The workflow is first designed for annotating points and is then thoroughly verified. Two major quality checks are 1) its ease of use for a large volunteer group, bearing different domain expertise, and 2) its fitness towards the key project deliverables. In our case, the workflow defines 'order' and 'name' for each facial point. Further, it also comprises a clear action-plan in case of ambiguities (e.g., invisible landmarks) by linking a professionally developed help page. It shows instructions and illustrations to annotate points across all possible species across diverse poses. Lastly, our workflow is thoroughly tested by a 5-member team of experts and it took 20 man-hours of effort.

\begin{SCfigure}[][!tp]
\centering
    \includegraphics[width=0.6\linewidth]{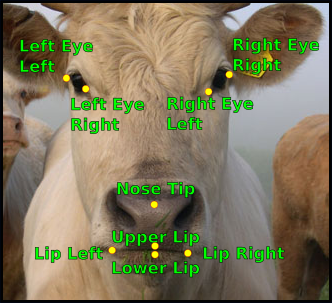}
    \hspace{-1em}
   \caption{\small Nine landmarks markup scheme used for annotation of faces in AnimalWeb. The markup scheme covers major facial features around key face components (eyes, nose, and lips) while keeping the total landmark count low.} 
   \label{fig:markup_scheme}
\end{SCfigure}

\noindent \textbf{9 pts. markup scheme.} The annotator interface in our case required annotators to adhere to the 9 landmarks markup scheme as shown in Fig.~\ref{fig:markup_scheme}. We believe that 9 landmarks provide good trade-off between annotation effort and facial features coverage.
\subsection{Facial Point Annotation}
\label{subsection:fp_annot}
After workflow development, the project is exposed to a big pool of Zooniverse volunteers for annotating facial landmarks. These volunteers have a prior experience of annotating many different successful citizen science projects related to animals. Every face is annotated by at least 5 different volunteers and this equals a labour-intensive effort of $\sim$5408 man-hours in total. Multiple annotations of a single face improves the likelihood of recovering annotated points closer to the actual location of facial landmarks, provided more than half of these multiple annotations qualify this assumption. To this end, we choose to take median value of multiple annotations of a single face.

The annotation portal allows annotators to raise a query with the experts throughout the annotation life cycle. This also helps in removing many different annotation ambiguities for other volunteers as well who might experience the same later in time. The whole exercise of Zooniverse crowdsourcing took 80 man-hours of experts’ time.



\subsection{Refining Annotations}
\label{subsection:ref_annot}

Annotations performed by zooniverse volunteers can be inaccurate and missing for some facial points. Further they could be inconsistent, and unordered. Unordered point annotations result if, for instance, left eye landmark is swapped with right eye. Above mentioned errors are in some sense justifiable since point annotations on animal faces, captured in real-world settings, is a complicated task.

We hired a small team of 4 trained volunteers for refinement. This team task was to perform manual corrections and it was supervised by an expert. The refinement completed in two passes listed below and took 438 man-hours of manual effort.

\noindent \textbf{First pass.}  In the first pass, major errors were rectified e.g., correcting points ordering. This refinement proceeded species-wise to enforce consistency in annotations
across every possible species in the dataset. A total of 548 man-hours were spent in the first pass.

\noindent \textbf{Second pass.} In the second pass, pixel perfect annotations were ensured by
cross-annotator review. For instance, the refinements on the portion of the dataset done by some member in the first pass is now reviewed and refined by another member of the team.


\section{Benchmarking AnimalWeb}
We extensively benchmark AnimalWeb for face alignment task. In addition, we demonstrate multi-task applications by demonstrating experimental results for two other related tasks: face detection and fine-grained image recognition. 


\subsection{Animal Facial Point Localization}
We select the state-of-the-art method in 2D human face alignment for evaluating the proposed dataset. Specifically, we take Hourglass (HG) deep learning based architecture; it has shown excellent results on a range of challenging 2D face alignment datasets \cite{bulat2017far,tang2018quantized} and competitions \cite{xiong2017combining}. 

\noindent \textbf{Datasets.} 300W-public, 300W-private, and COFW are deemed the most popular and challenging benchmarks for 2D human face alignment, and are publicly available. 300W-public contains 3148 training images and 689 testing images. 300W-private comprises 600 images for testing only. We only use COFW for testing purposes; its testing set contains 507 images.

\noindent \textbf{Evaluation Metric.} We use Normalized Mean Error (NME) as the face alignment evaluation metric,
\begin{equation*}
 \text{NME} = \frac{1}{N} \sum_{i=1}^{N}\sum_{l=1}^{L}(\frac{\parallel x{i}^{'}(l) - x{i}^{g}(l)\parallel }{d_{i}}).
\end{equation*}
It calculates the Euclidean distance between the predicted and the ground truth point locations and normalizes by $d_{i}$. We choose ground truth face bounding box size as $d_{i}$, as other measures such as Interocular distance could be biased for profile faces \cite{ramanan2012face}. In addition to NME, we report results using Cumulative Error Distribution (CED) curves, Area Under Curve (AUC) @0.08 (NME) error, and Failure Rate (FR) @0.08 (NME) error.

\noindent \textbf{Training Details.} For all our experiments, we use the settings described below to train HG networks both for human datasets and AnimalWeb. Note, these are similar settings as described in \cite{tang2018quantized,xiong2017combining} to obtain top performances on 2D face alignment datasets. We set the initial learning rate to $10^{-4}$ and used a mini-batch of 10. During the process, we divide the learning rate by 5, 2, and 2 at 30, 60, and 90 epochs, respectively, for training a total of 110 epochs. We also applied random augmentation: rotation (from -$30^{o}$ to $30^{o}$), color jittering, scale noise (from 0.75 to 1.25). All networks were trained using RMSprop \cite{tieleman2012divide}.

\begin{table*}[t]
\centering
\resizebox{0.80\textwidth}{!}{
\begin{tabular}{l|c| c |c| c}
\hline
\textbf{Datasets}   &\multicolumn{2}{c}{\textbf{9 pts.}}  &\multicolumn{2}{c}{\textbf{68 pts.}} \\
\cline{2-5}
&HG-2 &HG-3 &HG-2 &HG-3 \\
\hline
300W (common) &1.21/84.8/0.18 &1.19/85.0/0.00 &1.26/84.1/0.00 &1.25/84.2/0.00 \\
300W (full)  &1.42/82.1/0.14 &1.40/82.4/0.00 &1.41/82.2/0.00 &1.40/82.3/0.00\\
300W (challenging) &2.28/71.4/0.00 &2.25/71.7/0.00 &2.03/74.5/0.00 &2.01/74.8/0.00\\
300W (private) &2.26/72.2/0.66 &2.31/72.4/1.16 &1.82/77.5/0.50 &1.77/77.8/0.16 \\
COFW  &3.43/60.0/3.74 &3.26/61.3/3.55 &2.66/67.2/1.97 &2.60/68.2/1.57\\ 
AnimalWeb (Known) &5.35/47.4/17.2 &5.23/47.7/16.5 &- &- \\
AnimalWeb (Unknown) &6.50/39.6/23.8 &6.44/39.5/23.1 &- &- \\ 
\hline
\end{tabular}
}
\caption{\small Accuracy comparison between the AnimalWeb and 5 different human face alignment benchmarks when stacking 2 and 3 modules of HG network. We show human face alignment results both in terms of 68 pts. and 9 pts. 
Format for each table entry is: NME error/AUC@0.08 (NME) error/FailureRate@0.08 (NME) error. All results are in \%.}\label{tab:all_results_alignment}
\end{table*}

\begin{figure*}[htp]
\centering
    \includegraphics[clip=true, trim=0.5cm 0.2cm 0.5cm 0.5cm, width=0.42\linewidth]{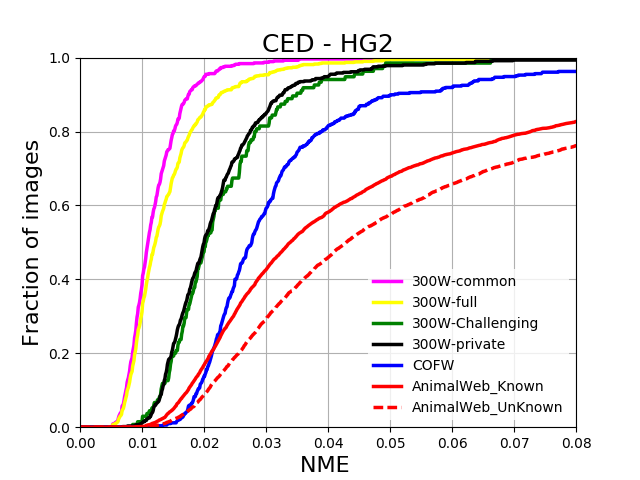}
    \includegraphics[clip=true, trim=0.1cm 0.2cm 1cm 0.5cm,width=0.42\linewidth]{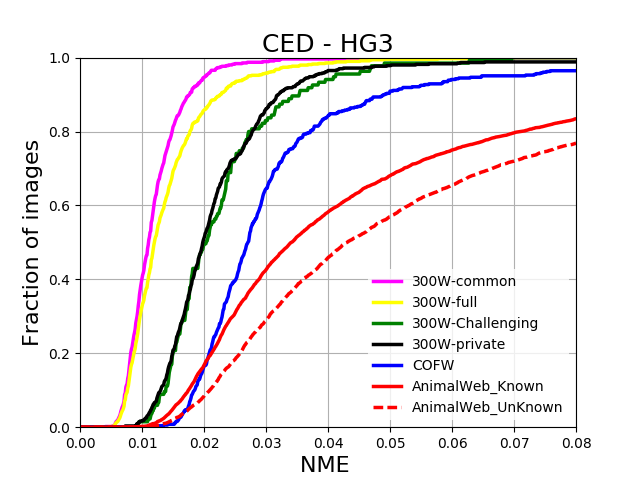}
   \caption{\small Comparison between AnimalWeb and popular face alignment datasets using HG-2\&3 networks. AnimalWeb results are reported for both Known and Unknown Species evaluation.}
   \label{fig:ced_hg}
\end{figure*}

AnimalWeb is assessed under two different train/test splits. The first setting randomly takes 80\% images for  training  and the rest 20\% for testing purposes from each specie. \footnote{For validation, we recommend using 10\% of the data from the training set.} We term this as `\textit{Known species evaluation}' since during training the network sees examples from every species expected upon testing phase. This setting can also be regarded as so-called `\textit{few-shot face alignment}'. 

The second setting randomly divides all species into 80\% for training and 20\% for testing. We term it as `\textit{Unknown species evaluation}' as the species encountered in testing phase are not available during training. This setting can also be deemed as so-called `\textit{zero-shot face Alignment}' (ZFA). Unknown species evaluation is, perhaps, more akin to real-world settings than its counterpart. This is because it is quite likely for a deployed facial behaviour monitoring system to experience some species that were unavailable at training. This setting is also more challenging compared to the first because facial appearance of species encountered during testing can be quite different to the ones available at training time.

\begin{figure*}[!htp]
\centering
\includegraphics[width = 0.95\linewidth]{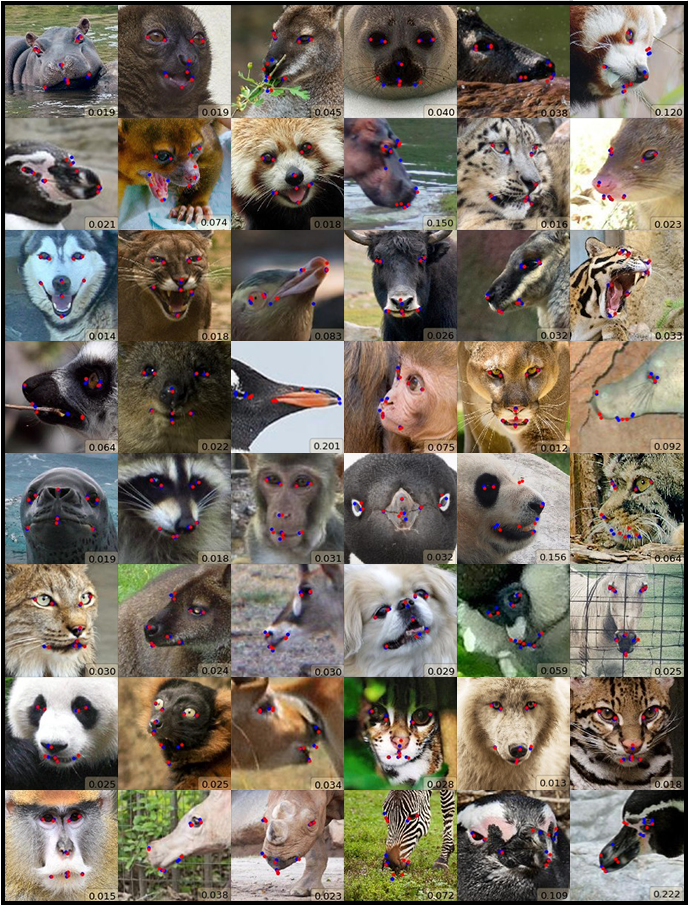} 
\caption{Example fittings from AnimalWeb under Known species evaluation. Red points denote fittings results of HG-3 and blue points are the ground truths.} 
\label{fig:known_species_first} 
\end{figure*}

\noindent \textbf{Known Species Evaluation.}
Tab.~\ref{tab:all_results_alignment} reveals comparison between AnimalWeb and 3 different human face alignment benchmarks, 300W-public, 300W-private, and COFW, when stacking 2 and 3 modules of HG network. Human face alignment results are shown both in terms of 68 pts. and 9 pts. To make fair comparison, the 9 pts. chosen on human faces are the same as for animal faces. Further, 9 pts. results correspond to the model trained with 9 pts. on human faces. We see a considerable gap (NME error difference) between all the results for human face alignment datasets and AnimalWeb. For instance, the NME error difference between COFW tested using HG-2 network is $\sim 1$ unit with AnimalWeb under the known species evaluation protocol. We observe a similar trend in the CED curves displayed in Fig.~\ref{fig:ced_hg}. Performance of COFW dataset, the most challenging among human faces, is 15\% higher across the whole spectrum of pt-pt-error. Finally, we display some example fittings under known species evaluation settings in Fig.~\ref{fig:known_species_first}. We see that the existing best method struggles under various in-the-wild situations exhibited in AnimalWeb.




\begin{figure*}[htp]
\begin{center}
    \includegraphics[width=1.0\linewidth]{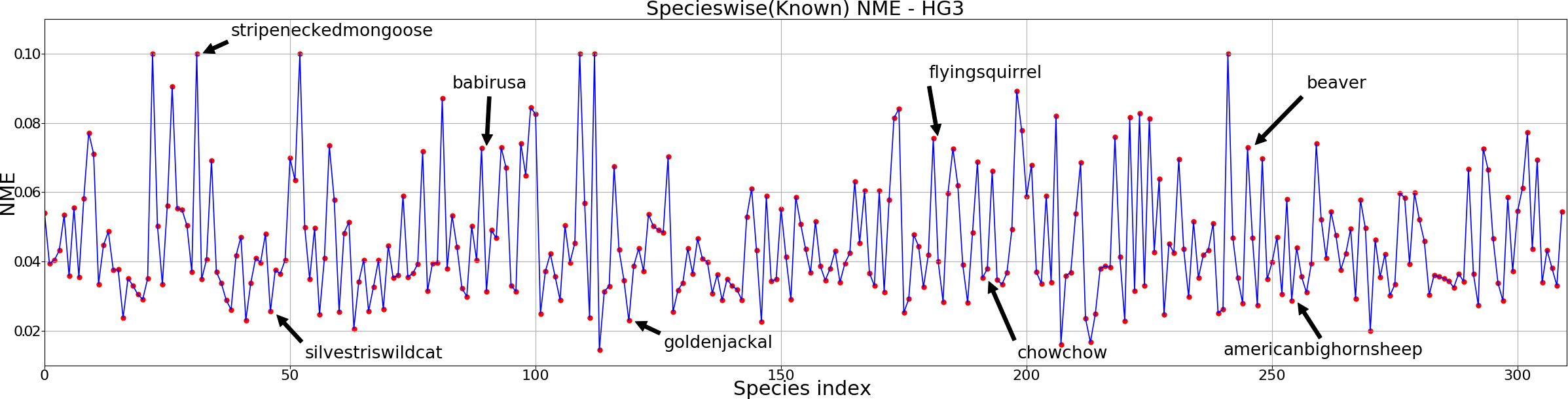}
\end{center} 
   \caption{\small Specie-wise results for AnimalWeb under Known Species settings. Y-axis indicates average NME for each specie.} 
   \label{fig:classwise}
\end{figure*}

\begin{figure}[ht]
  \centering
  \scalebox{1}{
  \includegraphics[width=1\columnwidth]{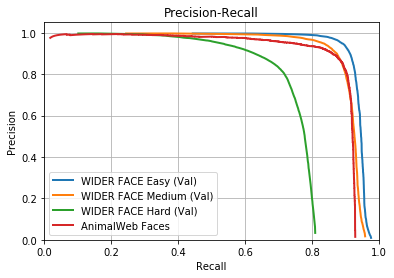} }
     \captionof{figure}{\small Precision-recall curve for AnimalWeb and WIDER Face datasets. }
    \label{fig:pr_cruve}
 \end{figure}

\begin{figure*}[!htp]
\centering
\includegraphics[width = 0.95\linewidth]{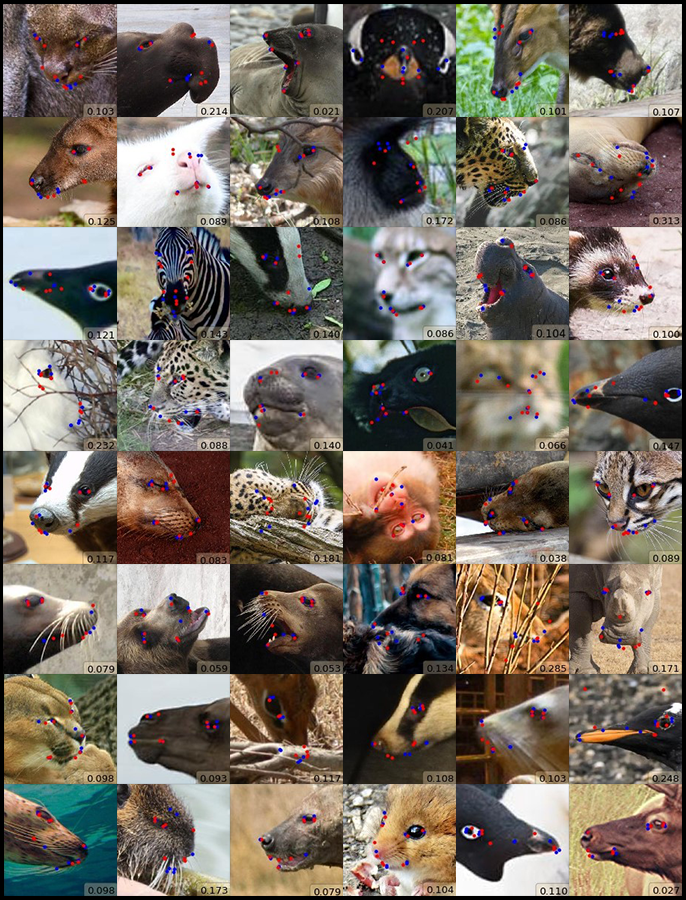} 
\caption{Example fittings from AnimalWeb under Unknown species evaluation. Red points denote fittings results of HG-3 and blue points are the ground truths.} 
\label{fig:unknown_species_first} 
\end{figure*}

Fig.~\ref{fig:classwise} depicts specie-wise testing results for AnimalWeb. For each specie, results are averaged along the number of instances present in it. We observe poorer performance for some species compared to others. This is possibly due to large intra-specie variations coupled with the scarcity of enough training instances relative to others. For instance, \textit{stripedneckedmongoose} species have only 8 training samples compared to \textit{silvesteriswildcat} species populated with 26 training examples. 

We report pose-wise results based on yaw angle in Tab.~\ref{tab:pose_wise}. It can be seen that AnimalWeb is challenging for large poses. The performance drops as we move towards the either end of (shown) yaw angle spectrum from $[-45^{o},45^{o}]$ range. Further, Tab.~\ref{tab:facesize_wise} shows results for AnimalWeb under different face sizes. We observe room for improvement across a wide range of face sizes.  

\noindent \textbf{Unknown Species Evaluation.}
Here, we report results under unknown species settings. Note, we randomly choose 80\% of the species for training and the rest 20\% for testing. Tab.~\ref{tab:all_results_alignment} draws comparison between unknown species settings and its counterpart. As expected, accuracy is lower for unknown case versus the known case. For example, HG-2 displays $\sim1$ unit poor performance under unknown case in comparison to known. Animal faces display much larger inter-species variations between some species. For example, \textit{adeliepenguins} and \textit{giantpandas} whom face appearances are radically different (see 5th row in Fig.~\ref{fig:known_species_first}). Fig.~\ref{fig:unknown_species_first} displays example fittings under this setting. We see that the fitting quality is low for a few frontal poses since the face appearance of species seen during training could be very different to species encountered when testing.

Low performance of existing face alignment algorithms under unknown species setting present obvious opportunities for the design and development of so-called 'zero-shot face alignment algorithms' that are robust to unseen facial appearance patterns. For instance, novel methods that can better leverage shared prior knowledge and similarities across seen species to perform satisfactorily under unknown species. 



\begin{table}[t]
\centering
\scalebox{0.9}{
\begin{tabular}{lccccc}
\hline
Yaw  &-90$^{o}$ &[-90$^{o}$,-45$^{o}$] &[-45$^{o}$,45$^{o}$] &[45$^{o}$,90$^{o}$] &90$^{o}$ \\ \hline
Faces &594  &877  &1226 &953    & 600    \\
NME &7.35 &5.02 &3.31 &5.50 &6.96 \\
\hline
\end{tabular}} 
\caption{\small Pose-wise NME(\%) based on yaw-angles with HG-3 under Known species settings of AnimalWeb.} 
\label{tab:pose_wise}
\end{table}

\begin{table}[t]
\centering
\scalebox{0.9}{
\begin{tabular}{lccc}
\hline
Face size  &[0,0.16] &[0.16,0.32] &[0.32,0.48] 
\\ \hline
Faces &3185 &911 &140 
\\
NME &5.45 &4.46 &5.19 
\\
\hline
\end{tabular}} 
\caption{\small NME(\%) w.r.t face size distribution with HG-3 under Known species settings of AnimalWeb. Face sizes are normalized by the corresponding image sizes.} 
\label{tab:facesize_wise}
\end{table}

\begin{figure*}[ht]
\begin{center}
    \includegraphics[width=0.96\linewidth, clip=true, trim=0cm 1cm 0cm 0cm]{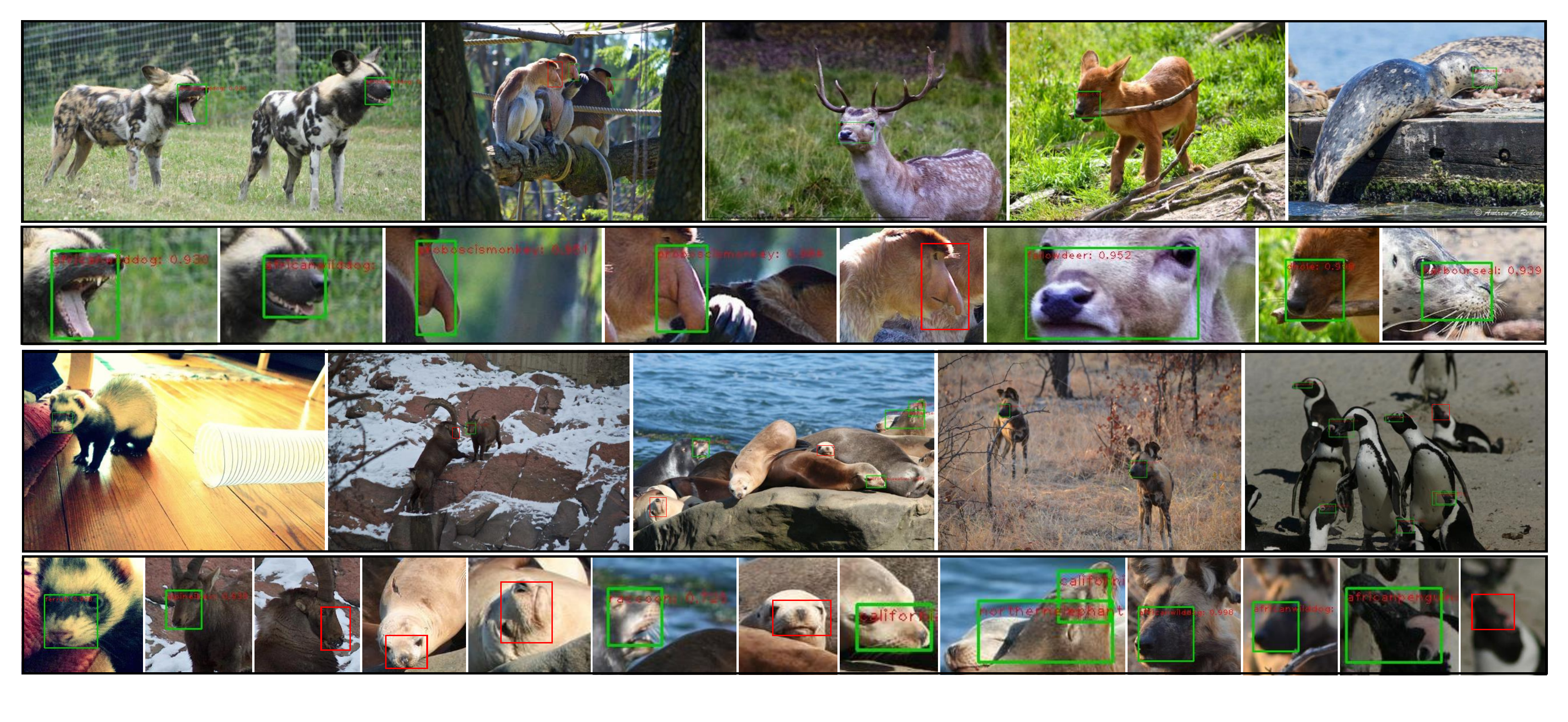}
\end{center} 
   \caption{\small Example face detections from AnimalWeb. Green/red boxes denote true/missed detections from Faster-RCNN \cite{ren2015faster} baseline.} 
   \label{fig:qual_det_results}
\end{figure*}


\subsection{Animal Face Detection} We evaluate the performance of animal face detection using a Faster R-CNN \cite{ren2015faster} baseline. Our ground-truth is a tightly enclosed face bounding box for each animal face, that is obtained by fitting the annotated facial landmarks. We first evaluate our performance on the face localization task. We compare our dataset with one of the most challenging human face detection dataset WIDER Face \cite{yang2016wider} in terms of Precision-Recall curve (Fig.~\ref{fig:pr_cruve}). Note that WIDER Face is a large-scale dataset with $393,703$ face instances in 32K images and introduces three protocols for evaluation namely `easy', `medium' and `hard' with the increasing level of difficulty. The performance on our dataset lies close to that of medium curve of WIDER Face, which shows that there exists a reasonable margin of improvement for animal face detection. We also compute overall class-wise detection scores where the Faster R-CNN model achieves a mAP of 0.636. Some qualitative examples of our animal face detector are shown in Fig.~\ref{fig:qual_det_results}.

\subsection{Fine-grained species recognition}
Since our dataset is labeled with fine-grained species, one supplementary task of interest is the fine-grained classification. We evaluate the recognition performance on our dataset by applying Residual Networks \cite{he2016deep} with varying depths (18, 34, 50 and 101). Results are reported in Tab.~\ref{tab:fine-grained_re}. We can observe a gradual boost in top-1 accuracy as the network capacity is increased. Our dataset shows a similar difficulty level in comparison to other fine-grained datasets of comparable scale, e.g., CUB-200-2011 \cite{dataset_cub} and Stanford Dogs \cite{dataset_dogs} with 200 and 120 classes, respectively. A ResNet50 baseline on CUB-200 and Stanford Dogs achieve an accuracy of 81.7\% and 81.1\% \cite{sun2018multi}, while the same network achieves an accuracy of 80.04\% on AnimalWeb. 

\begin{table}[t]
\centering
\scalebox{0.9}{
\begin{tabular}{lcccc}
\hline
Network   &ResNet18 &ResNet34 &ResNet50 &ResNet101    \\ \hline
Accuracy           &76.49     &79.22     &80.04     &81.06    \\
\hline
\end{tabular}} 
\caption{\small Fine-grained recognition accuracy on AnimalWeb. Top-1 accuracies (in \%) are reported using four ResNet variants \cite{he2016deep}.} 
\label{tab:fine-grained_re}
\end{table}

\section{Conclusion}
In this paper, we introduce a large-scale, hierarchical dataset, named AnimalWeb, of annotated animal faces. It features 21.9K faces from 334 diverse animal species while exploring 21 different orders across animal biological taxonomy. Each face is consistently annotated with 9 fiducial landmarks centered around key facial components. It is structured and scalable by design. Benchmarking AnimalWeb under new settings for face alignment, employing current state-of-the-art method, reveal its challenging nature. It conjectures that existing best methods for (human) face alignment are suboptimal for this task, highlighting the need for specialized and robust algorithms to analyze animal faces. We also show the applications of the dataset for related tasks, specifically face detection and fine-grained recognition. Results conclude that the proposed dataset is a good experimental foundation for algorithmic advances in CV and the resulting technology for the betterment of society and economy.


%
%

\bibliographystyle{spmpsci}      
\bibliography{egbib}   


\end{document}